\documentclass{article}
\usepackage{fullpage}
\usepackage{euler}
\usepackage{amsmath}
\usepackage{proof}
\usepackage{mathabx}
\usepackage{color}
\usepackage{graphicx}
\usepackage[shortlabels]{enumitem}

\newcommand{\lang}{\ensuremath{\mathcal{L}}}
\newcommand{\To}{\ensuremath{\Rightarrow}}
\newcommand{\tTo}{\ensuremath{\rightsquigarrow}}
\newcommand{\Rules}{\ensuremath{\mathcal{R}}}
\newcommand{\mc}[1]{\ensuremath{\mathcal{#1}}}
\newcommand{\conc}{\ensuremath{\mathtt{Conc}}}
\newcommand{\sub}{\ensuremath{\mathtt{Sub}}}
\newcommand{\prem}{\ensuremath{\mathtt{Prem}}}
\newcommand{\toprule}{\ensuremath{\mathtt{TopRule}}}
\newcommand{\rules}{\ensuremath{\mathtt{Rules}}}
\newcommand{\args}[1]{\ensuremath{\mathcal{A}({#1})}}
\newcommand{\closure}{\ensuremath{Cl}}

\newcommand{\acro}[1]{\textsc{\lowercase{#1}}}

\newcommand{\aspic}{\textsc{aspic$^\mathsf{+}$}}

\newtheorem{definition}{Definition}
\newtheorem{proposition}{Proposition}

\newtheorem{example}{Example}
\newenvironment{proof}{\vskip 10pt\noindent\textbf{Proof:}}{\hfill$\Box$}

\newcommand{\sdp}[1]{\textsf{{\color{blue} #1}}}
\newcommand{\ac}[1]{\textsf{{\color{red} #1}}}

\title{\textsf{Two forms of minimality in \aspic}}
\author{Zimi Li\\
	Department of Computer Science\\ 
	Graduate Center, City University of New York\\
	E-mail: zli2@gradcenter.cuny.edu 
	\and 
	Andrea Cohen\\
	Institute for Computer Science and Engineering\\
	UNS-CONICET\\
	E-mail: ac@cs.uns.edu.ar\\
	\and Simon Parsons\\
	Department of Informatics\\
	King's College London\\
	E-mail: simon.parsons@kcl.ac.uk}

\begin{document}

\maketitle

\begin{abstract}
  Many systems of structured argumentation explicitly require that the facts and rules that make up the argument for a conclusion be the minimal set required to derive the conclusion.
  \aspic\ does not place such a requirement on arguments, instead requiring that every rule and fact that are part of an argument be used in its construction.
  Thus \aspic\ arguments are minimal in the sense that removing any element of the argument would lead to a structure that is not an argument.
  In this brief note we discuss these two types of minimality and show how the first kind of minimality can, if desired, be recovered in \aspic.
\end{abstract}

\section{Introduction}
\label{sec:intro}

A large part of the work on computational argumentation is concerned with \emph{structured}, or \emph{logic-based} argumentation.
  In this work, much of the focus is on the way that arguments are constructed from some set of components, expressed in some logic.
  At this point, perhaps the most widely studied system of structured argumentation is \aspic, which builds on what is now quite a lengthy tradition, a tradition which goes back at least as far as \cite{pollock:cognitive}.
  In addition to Pollock's work on \acro{oscar} \cite{pollock:carpentry,pollock:jancl}, we can count the work of Loui \cite{loui:ci}, Krause \emph{et al.} \cite{krause:ci}, Prakken and Sartor \cite{prakken:sartor:jancl}, Besnard and Hunter \cite{besnard:hunter:aij}, Amgoud and Cayrol \cite{amgoud:cayrol:amai}, Garc\'ia and Simari \cite{garcia:simari:tplp} and Dung \emph{et al.} \cite{dung:kowalski:toni:aij} as being in the same lineage.
  \aspic\ \cite{modgil:prakken:ai,prakken:ac} is more recent, but very influential, providing a very general notion of argumentation that captures many of the structured systems which predate it.
  In all these systems, there is, often explicitly, a notion of an argument as a pair $\langle c, \Delta\rangle$ which relates the conclusion of the argument, $c$, and the set of statements $\Delta$ from which that conclusion is derived.
  The form of derivation, and what these ``statements'' consist of, are two of the aspects of these systems which vary widely.

One difference between \aspic\ and other systems of structured argumentation is that many of the latter require that arguments be minimal in the sense that the set $\Delta$ in any argument $\langle c, \Delta\rangle$ has to be minimal.
  That is, $\Delta$ has to be the smallest set from which $c$ can be derived.
  We can find this explicitly expressed, for example, in \cite{amgoud:cayrol:amai,besnard:hunter:aij,garcia:simari:tplp}.
  In contrast, like the assumption-based system from \cite{dung:kowalski:toni:aij}, \aspic\ does not explicitly require arguments to be minimal in this sense.
  Instead \aspic\ arguments satisfy a different form of minimality in which arguments cannot include premises or rules that are not used in the derivation of their conclusion.
  In recent work using \aspic\cite{cohen:parsons:sklar:mcburney:support}, we discovered some cases in which the difference between these two forms of minimality was important, and so needed to investigate those differences in the context of \aspic.
  In this note we report our findings.

Note that while the first form of minimality is stronger than the native minimality of \aspic, because there are \aspic\ arguments that are not minimal in this sense, this form of minimality is completely compatible with \aspic, and indeed with assumption-based argumentation (which shares the same mechanism for defining an argument).
  As we show, when the stronger form of minimality is required, we can simply invoke a definition for arguments in \aspic\ which does require this form of minimality.
  


\section{Background}
\label{sec:background}

\aspic\ is deliberately defined in a rather abstract way, as a system
with a minimal set of features that can capture the notion of
argumentation. This is done with the intention that it can be
instantiated by a number of concrete systems that then inherit all of
the properties of the more abstract system. \aspic\ starts from a
logical language $\lang$ with a notion of negation. A given
instantiation will then be equipped with inference rules, and
\aspic\ distinguishes two kinds of inference rules: strict rules and
defeasible rules. Strict rules, denoted using $\to$, are rules whose
conclusions hold without exception. Defeasible rules, denoted $\To$,
are rules whose conclusions hold unless there is an exception.

The language and the set of rules are used to define an \emph{argumentation system}:
\begin{definition}[Argumentation System]
An \emph{argumentation system} is a tuple $AS = \langle \lang, \overline{\phantom{\cdot}}, \Rules, n \rangle$ where: 
\begin{itemize}
\item $\lang$ is a logical language.
\item $\overline{\phantom{\cdot}}$ is a function from $\lang$ to $2^\lang$ , 
      such that:
      \begin{itemize}
      \item $\varphi$ is a \emph{contrary} of $\psi$ if $\varphi\in\overline{\psi}$, $\psi\not\in\overline{\varphi}$;
      \item $\varphi$ is a \emph{contradictory} of $\psi$ if $\varphi\in\overline{\psi}$, $\psi\in\overline{\varphi}$;
      \item each $\varphi\in\lang$ has at least one contradictory.
      \end{itemize}
\item $\Rules = \Rules_s \cup \Rules_d$ is a set of strict ($\Rules_s$) and defeasible ($\Rules_d$) inference rules of the form $\phi_1, \ldots, \phi_n \to \phi$ and $\phi_1, \ldots, \phi_n \To \phi$ respectively (where $\phi_i, \phi$ are meta-variables ranging over wff in $\lang$), and $\Rules_s \cap \Rules_d = \emptyset$.
\item $n : \Rules_d \mapsto \lang$ is a naming convention for defeasible rules.
\end{itemize}
\end{definition}
The function $\overline{\phantom{\cdot}}$ generalizes the usual symmetric notion of negation to allow non-symmetric conflict between elements of $\lang$. The contradictory of some $\varphi\in\lang$ is close to the usual notion of negation, and we denote that $\varphi$ is a \emph{contradictory} of $\psi$ by ``$\varphi = \neg\psi$''.
%
The naming convention for defeasible rules is necessary because \aspic\ allows for rules that deny the applicability of other rules. Giving rules names makes this easy to accomplish.

An argumentation system, as defined above, is just a language and some rules which can be applied to formulae in that language. To provide a framework in which reasoning can happen, we need to add information that is known, or believed, to be true. In \aspic\ this information makes up a \emph{knowledge base}:
\begin{definition}[Knowledge Base]
A  \emph{knowledge base} in an argumentation system $\langle \lang, \overline{\phantom{\cdot}}, \Rules, n \rangle$ is a set $\mc{K} \subseteq \lang$ consisting of two disjoint subsets $\mc{K}_n$ and $\mc{K}_p$.
\end{definition}
We call $\mc{K}_n$ the \emph{axioms} and $\mc{K}_p$ the \emph{ordinary premises}. We make this distinction between the elements of the knowledge base for the same reason that we make the distinction between strict and defeasible rules. We are distinguishing between those elements --- axioms and strict rules --- which are definitely true and allow truth-preserving inferences to be made, and those elements --- ordinary premises and defeasible rules --- which can be disputed. 

Combining the notions of argumentation system and knowledge base gives us the notion of an \emph{argumentation theory}:
\begin{definition}[Argumentation Theory]
An \emph{argumentation theory} $AT$ is a pair $\langle AS, \mc{K} \rangle$ of an argumentation system $AS$ and a knowledge base $\mc{K}$.
\end{definition}
We are now nearly ready to define an argument. But first we need to
introduce some notions which can be defined just understanding that an
argument is made up of some subset of the knowledge base $\mc{K}$, along
with a sequence of rules, that lead to a conclusion. Given this,
$\prem(\cdot)$\ returns all the premises, $\conc(\cdot)$\ returns the
conclusion and $\toprule(\cdot)$\ returns the last rule in the
argument. $\sub(\cdot)$\ returns all the sub-arguments of a given
argument, that is all the arguments that are contained in the given
argument.
%
\begin{definition}[Argument]
\label{def:argument}
An \emph{argument} $A$ from an argumentation theory $AT = \langle \lang, \overline{\phantom{\cdot}}, \rules, n \rangle, \mc{K}\rangle$ is:
\begin{enumerate}
\item $\phi$ if $\phi \in \mc{K}$ with: $\prem(A) = \{\phi\}$; $\conc(A) = \{\phi\}$; $\sub(A) = \{A\}$; $\toprule(A) = $ undefined.

\item $A_1, \ldots, A_n \to\phi$ if $A_i$, $1 \leq i \leq n$, are arguments and there exists a strict rule of the form \linebreak \mbox{$\conc(A_1), \ldots, \allowbreak\conc(A_n) \to\phi$} in $\Rules_s$.
$\prem(A) = \prem(A_1) \cup \ldots \cup \prem(A_n)$; $\conc(A) = \phi$; \allowbreak
$\sub(A) = \sub(A_1) \cup \ldots \cup \sub(A_n) \cup \{A\}; \toprule(A) = \conc(A_1), \ldots, \conc(A_n) \to\phi$.

\item $A_1, \ldots, A_n \To \phi$ if $A_i$, $1 \leq i \leq n$, are arguments and there exists a defeasible rule of the form \mbox{$\conc(A_1), \ldots, \allowbreak\conc(A_n) \To \phi$} in $\Rules_d$.
$\prem(A) = \prem(A_1) \cup \ldots \cup \prem(A_n)$; 
\mbox{$\conc(A) = \phi$}; $\sub(A) = \sub(A_1) \cup \ldots \cup \sub(A_n) \cup \{A\}; \toprule(A) = \conc(A_1), \ldots, \conc(A_n) \To \phi$.
\end{enumerate}
We write \args{AT} to denote the set of arguments from the theory $AT$.
\end{definition}
%
In other words, an argument is either an element of $\mc{K}$, or it is made up of a rule and its conclusion where each premise of the rule is the conclusion of a sub-argument.
From here on, we will use the symbol $\tTo$ when we do not care about distinguishing whether an argument uses a strict rule $\to$ or a defeasible rule $\To$. Thus, if we are making a statement about an argument $A = [B \tTo a]$, then we are making a statement about both arguments $A' = [B \to a]$ and $A'' = [B \To a]$. Similarly, when referring to a rule $a \tTo b$ we are referring to either a strict rule $a \to b$ or a defeasible rule $a \To b$, and in a context in which it does not matter whether the rule is strict or defeasible.

The above is a standard presentation of an argument in \aspic.
In this report we wish to refer to an additional element of an argument, and to describe an argument in somewhat different way.
In particular, we wish to refer to $\rules(A)$, which identifies all the strict or defeasible rules used in the argument $A$.

\begin{definition}[Argument Rules]
Let $AT = \langle AS, \mc{K} \rangle$ be an argumentation theory and $A \in \args{AT}$. We define the set of rules of $A$ as follows:
\[
\rules(A) =
\begin{cases} 
\emptyset &  A \in \mc{K}\\ \\
\{\toprule(A)\} \cup \bigcup_{i=1}^{n} \rules(A_i) &  A = A_1, \ldots, A_n \tTo \conc(A)
\end{cases}
\]
\end{definition}
We can then describe an argument $A$ as a triple:
\[
(G, R, p)
\]
where $G = \prem(A)$ are the \emph{grounds} on which $A$ is based, $R = \rules(A)$ is the set of rules that are used to construct $A$ from $G$, and $p= \conc(A)$ is the conclusion of $A$.

%
%

\begin{example}\label{ex:arguments}
Consider that we have an argumentation system $AS_{\ref{ex:arguments}} = \langle \lang_{\ref{ex:arguments}}, \overline{\phantom{\cdot}}, \Rules_{\ref{ex:arguments}}, n \rangle$, where $\lang_{\ref{ex:arguments}} = \{p,q,r,s,t,u, \linebreak v, \neg p, \neg q, \neg r, \neg s, \neg t, \neg u, \neg v\}$, $\Rules_{\ref{ex:arguments}} = \{p,q\tTo r; t,u \tTo r; r\tTo s; u\tTo v\}$. By adding the knowledge base $\mc{K}_{\ref{ex:arguments}} = \{p, q, t, u\}$ we obtain the argumentation theory $AT_{\ref{ex:arguments}} = \langle AS_{\ref{ex:arguments}}, \mc{K}_{\ref{ex:arguments}}\rangle$, from which we can construct the following arguments:
\[
\begin{array}{llll}
A_1 = [p]; & A_2 = [q]; & A_3 = [A_1, A_2 \tTo r]; & A = [A_3 \tTo s];\\
B_1 = [t]; & B_2 = [u]; & B_3 = [B_1, B_2 \tTo r]; & B = [B_3 \tTo s]
\end{array}
\]
such that 
$A_1 = (\{p\}, \emptyset, p)$, 
$A_2 = (\{q\}, \emptyset, q)$,
$A_3 = (\{p, q\}, \{p, q \tTo r\}, r)$,
$A = (\{p, q\}, \{p, q \tTo r; \linebreak r \tTo s\}, s)$,
$B_1 = (\{t\}, \emptyset, t)$,
$B_2 = (\{u\}, \emptyset, u)$,
$B_3 = (\{t,u\}, \{t, u \tTo r\}, r)$ 
and $B = (\{t, u\}, \{t, u \tTo r; \linebreak r \tTo s\}, s)$.
\end{example}



\section{Minimality}
\label{sec:minimal}

Now, as mentioned above, unlike some definitions of arguments in the literature---for example \cite{amgoud:cayrol:amai,garcia:simari:tplp}---Definition~\ref{def:argument} does not mention the minimality of the grounds or the set of rules.
This does not mean that \aspic\ arguments are not, in some sense, minimal, as we we will now show.

The following example illustrates the fact that any element (proposition or rule) in the grounds and rules of an argument needs to be used in the derivation of the conclusion of that argument:
\begin{example}\label{ex:relevance:grounds-rules}
Given the argumentation theory from Example~\ref{ex:arguments}, the structure
\[ C = (\{p, q, t, u\}, \{p, q \tTo r; t, u \tTo r; r \tTo s\}, s) \]
is not an argument.
In particular, $C$ is not an argument because the third clause of Definition~\ref{def:argument} only justifies adding the rules and grounds of one argument for each premise of the rule that is the subject of the clause. Thus, it allows $p, q \tTo r$ to be added to an argument with conclusion $r$, or it allows $t, u \tTo r$ to be added, but it does not permit both to be added.
Similarly, 
\[
	\begin{array}{l}
	D = (\{p, q\}, \{p, q \tTo r; r \tTo s; u \tTo v\}, s)\\
	E = (\{t, u\}, \{t, u \tTo r; r \tTo s; u \tTo v\}, s)
	\end{array}
\]
are not arguments because Definition~\ref{def:argument} does not allow rules that are not used in the derivation of the conclusion of an argument to be part of the set of rules of that argument. 
Finally, neither of
\[
	\begin{array}{l}
	F = (\{p, q, t,u\}, \{p, q \tTo r; r \tTo s\}, s)\\
	G = (\{p, q, t, u\}, \{t, u \tTo r; r \tTo s\}, s)
	\end{array}
\]
are arguments, because Definition~\ref{def:argument} does not allow the addition of propositions to the grounds of an argument if they do not correspond to premises of a rule in the argument.
\end{example}

Thus, as the preceding example shows, an argument $A = (G, R, p)$ cannot contain any elements in $G$ or $R$ that are not used in the derivation of $p$, and so Definition~\ref{def:argument} implies that arguments are minimal in the sense that they do not contain any extraneous propositions or rules. This intuition is also pointed out by the authors in~\cite{modgil:prakken:ai}, and we formalize it in the following proposition:
\begin{proposition}\label{prop:relevance:grounds-rules}
Let $AT = \langle AS, \mc{K}\rangle$ be an argumentation theory and $A \in \args{AT}$.
It holds that either:
\begin{enumerate}[(a)]
	\item $A = (\{p\}, \emptyset, p)$; or
	\item $A = (G, R, p)$ and 
		\begin{enumerate}[i.]
			\item for every $g \in G$: there exists $A' \in \sub(A)$ such that $A' = (\{g\}, \emptyset, g)$ and there exists $r \in R$ such that $r = p_1, \ldots, g, \ldots,  p_n \tTo p'$; and
			\item for every $r' \in R$ such that $r' = p_1, \ldots, p_m \tTo p''$: there exists $A'' \in \sub(A)$ such that $A'' = (G'', R'' \cup \{r'\}, p'')$.
		\end{enumerate} 
\end{enumerate}
\begin{proof}
Definition~\ref{def:argument} includes three clauses that define when $A = (G, R, p)$ is an argument. In the first clause, the base case of the recursive definition, $p \in \mc{K}$, $R$ is empty and $G = \{p\}$, satisfying case (a).

The rest of this proof concerns case (b). Now, the second and third clauses of Definition~\ref{def:argument}, which define the recursive step of the definition, tells us that $(G, R, p)$ is an argument if there exists a rule $r \in R$ such that $r = p_1, \ldots, p_n \tTo p$ and for each $p_i$ $(1 \leq i \leq n)$ there exists an argument $A_i \in \args{AT}$ such that $\conc(A_i) = p_i$. In other words, for every premise $p_i$ of the rule $r$ there is a sub-argument $A_i$ of $A$ whose conclusion is that premise.
Unwinding each of those sub-arguments in turn, they are either of the form $(\{p_i\}, \emptyset, p_i)$, or can be deconstructed into a rule with sub-arguments for each premise, where that rule is in $R$. 
In the first of these cases, the first clause of Definition~\ref{def:argument} tells us that $p_i \in G$, and so case (b.i) holds. 
From the second of these cases we can infer that for every rule $p_1, \ldots, p_m\tTo p''\in R$, there is a sub-argument $(G'', R''\cup\{p_1, \ldots, p_m\tTo p''\}, p'')$ of $A$, and case (b.ii) is proved. 
\end{proof}
\label{prop:derivation}
\end{proposition}
In other words, given an argument $A = (G, R, p)$, Proposition~\ref{prop:relevance:grounds-rules} states that every element of the grounds $G$ and the rules $R$ is part of the derivation of $p$.
However, as the following example shows, Definition~\ref{def:argument} does not imply that for any argument $(G, R, p)$ there is no argument $(G', R', p)$ such that $G'\subset G$ and $R'\subset R$:
\begin{example}\label{ex:circularity-redundancy}
Consider that we have an argumentation system $AS_{1} = \langle \lang, \overline{\phantom{\cdot}}, \Rules_{1}, n \rangle$, where
$\lang = \{p, q, r, s, t, \linebreak \neg p, \neg q, \neg r, \neg s, \neg t\}$ 
and $\Rules_{1} = \{p,q\tTo s;  s \tTo q; q,r \tTo t\}$. 
By adding the knowledge base $\mc{K}_{1} = \{p, q, r\}$ we obtain the argumentation theory $AT_{1} = \langle AS_{1}, \mc{K}_{1}\rangle$, from which we can construct the following arguments:
\[
\begin{array}{l}
A_1 = [p];  A_2 = [q];  A_3 = [A_1, A_2 \tTo s];  A_4 = [A_3 \tTo q];  A_5 = [r];  
A = [A_4, A_5 \tTo t];\\ 
B = [A_2, A_5 \tTo t]
\end{array}
\]
such that $A = (\{p, q, r\}, \{p, q \tTo s; s \tTo q; q, r \tTo t\}, t)$ and $B = (\{q, r\}, \{q, r \tTo t\}, t)$. Here, it is clear that the grounds and rules of $B$ are a subset of those of $A$.

Consider now the set of rules $\Rules_{2} = \{p, q \tTo r; r \tTo s; s \tTo t; t \tTo r\}$. We can obtain a new argumentation system $AS_{2} = \langle \lang, \overline{\phantom{\cdot}}, \Rules_{2}, n \rangle$ and combine it with the knowledge base $\mc{K}_{1}$ to obtain the argumentation theory $AT_{2} = \langle AS_{2}, \mc{K}_{1}\rangle$, from which we can construct the arguments:
\[
\begin{array}{l}
C_1 = [p];  C_2 = [q];  C_3 = [C_1, C_2 \tTo r];  C_4 = [C_3 \tTo s];  C_5 = [C_4 \tTo t];  C = [C_5 \tTo r];\\ 
D = C_3 = [C_1, C_2 \tTo r]
\end{array}
\]
Here, $C = (\{p, q\}, \{p, q \tTo r; r \tTo s; s \tTo t; t \tTo r\}, r)$ and $D = (\{p, q\}, \{p, q \tTo r\})$ and thus, $\rules(D)$ is a subset of $\rules(C)$.

Finally, if we consider a set of rules $\Rules_{3} = \{p \tTo r; r \tTo s; q \tTo r; r, s \tTo t\}$ and a knowledge base $\mc{K}_{3} = \{p, q\}$ we can define an argumentation system $AS_{3} = \langle \lang, \overline{\phantom{\cdot}}, \Rules_{3}, n \rangle$ and an argumentation theory $AT_{3} = \langle AS_{3}, \mc{K}_{3}\rangle$, from which we obtain:
\[
\begin{array}{l}
E_1 = [p];  E_2 = [E_1 \tTo r];  E_3 = [E_2 \tTo s];  E_4 = [q]; E_5 = [E_4 \tTo r];  E = [E_5, E_3 \tTo t];\\ 
F = [E_2, E_3 \tTo t]
\end{array}
\]
In this case, $E = (\{p, q\}, \{p \tTo r; r \tTo s; q \tTo r; r, s \tTo t\}, t)$ and $F = (\{p\}, \{p \tTo r; r \tTo s; r, s \tTo t\}, t)$. As a result, the grounds and rules of $F$ are a subset of those of $E$.
\end{example}

At first sight, this seems a bit
contradictory. Example~\ref{ex:relevance:grounds-rules} and
Proposition~\ref{prop:derivation} show that arguments only contain
elements that are used in the derivation of their conclusion, yet
Example~\ref{ex:circularity-redundancy} shows that elements can be
removed from the grounds or the rules of an argument, and what remains
is still an argument. There is, however, no contradiction. Rather,
there are two ways in which this phnomenon might arise. The first is
illustrated by the first two cases in
Example~\ref{ex:circularity-redundancy}. Here we have arguments that
are \emph{circular}\footnote{We use the term ``circular'' to reflect
  the idea of circular reasoning \cite{walton:plausible} and ``begging
  the question'' \cite{sinnott:ajp}.}  --- if you follow the chain of
reasoning from premises to conclusion in $A$ in the above example, we
start with $q$, then derive $q$, then use $q$ to derive the final
conclusion; similarly, when considering $C$, we start with $p$ and $q$
to derive $r$, then derive $s$ and $t$ to derive (again) $r$. In $B$
and $D$ these loops are removed to give us more compact arguments with
the same conclusions.  The second way in which this phenomenon might
arise is illustrated by the third case in
Example~\ref{ex:circularity-redundancy}, where we have arguments that
are \emph{redundant}. Here the cause is the fact that the set of rules
provides two ways to derive $r$, one that relies on $p$ and another
that relies on $q$, and $r$ appears twice in the derivation of $t$,
once to produce $s$, and once when the rule $r, s\tTo t$ is
applied. $E$, the redundant argument, uses both of the rules for
deriving $r$ while $F$ uses just one of them, again providing a more
compact derivation.

Furthermore, as shown by the following example, circularity in arguments may lead to having two distinct arguments $A$ and $B$ such that their descriptions $(G, R, p)$ coincide. Hence, while we can extract a unique description $(G, R, p)$ from a given \aspic\ argument $A$, the reverse is not true.\footnote{This is a version of the issue pointed out by \cite[p119]{dung:kowalski:toni:aij}, that any inference-based description of an argument allows multiple arguments to be described in the same way. In fact what we have here is a stronger version of the problem, because \cite{dung:kowalski:toni:aij} pointed out the problem for arguments which, in our terms, were described just by their grounds and conclusion. What we have here is the problem arising even when we state the inference rules as well. This issue the is converse of the problem that describing arguments by their entire structure, as \aspic\ and the assumption-based argumentation of \cite{dung:kowalski:toni:aij} do, allows for redundant elements in the arguments, as we have just shown.}

\begin{example}\label{ex:circularity}
	Consider that we have an argumentation system $AS = \langle \lang, \overline{\phantom{\cdot}}, \Rules, n \rangle$, where
	$\lang = \{a, b, c, \neg a, \neg b, \neg c\}$ 
	and $\Rules = \{a \tTo c; c \tTo b; b \tTo a\}$. 
	We then add the knowledge base $\mc{K} = \{a\}$ to get the argumentation theory $AT = \langle AS, \mc{K}\rangle$. From this we can construct the following arguments:
	\[
	\begin{array}{llll}
	A_1 = [a]; & A_2 = [A_1 \tTo c]; & A_3 = [A_2 \tTo b]; & A = [A_3 \tTo a];\\
	B_1 = [A \tTo c]; & B_2=[B_1 \tTo b]; & B = [B_2 \tTo a]\\
	\end{array}
	\]
	Here, both arguments $A$ and $B$ are described by the triple $(G, R, a)$, where $G = \prem(A) = \prem(B) = \{a\}$, $R = \rules(A) = \rules(B) = \Rules$ and $a = \conc(A) = \conc(B)$.
\end{example}


Given the preceding analysis we can note that, even though the characterization of \aspic arguments accounts for some form of minimality (see~\cite{modgil:prakken:ai}), it allows for circular and redundant arguments.
These notions of circularity and redundancy are formalized next.
\begin{definition}\label{def:circular:argument}
Let $AT$ be an argumentation theory and $A \in \args{AT}$. 
We say that $A$ is a \emph{circular argument} if $\exists A_1, A_2 \in \sub(A)$ such that $A_1 \neq A_2$, $\conc(A_1) = \conc(A_2)$ and $A_1 \in \sub(A_2)$.
\end{definition}
Note that the usual definition of a circular argument in the literature \cite{sinnott:ajp,walton:plausible} involves starting with some premise and then inferring that premise --- a typical pattern is ``Assume $a$, then $a$ is true''. 
What we define here as circular is more general. 
\begin{example}\label{ex:circularity-twoforms}
Considering Example~\ref{ex:circularity-redundancy} in the light of
Definition~\ref{def:circular:argument} and looking at $A$, the two
sub-arguments that define its circularity are 
$A_1 = (\{q\}, \emptyset, q)$ and
$A_4 = (\{p, q\}, \{p, q\tTo s, s\tTo q\}, q)$.
Then, if we consider argument $C$, the two sub-arguments that define its circularity are	
$C_3 = (\{p, q\}, \{p, q \tTo r\}, r)$
and
$C = (\{p, q\}, \{p, q\tTo r, r\tTo s, s \tTo t, t \tTo r\}, r)$.
Here, $A$ follows the classic form of a circular argument. 
In contrast, $C$ illustrates a form of circularity not related to the premises of the argument.
\end{example}
Next, we formalize the notion of redundancy:
\begin{definition}\label{def:redundant:argument}
  Let $AT$ be an argumentation theory and $A \in \args{AT}$.
  We say that $A$ is a \emph{redundant argument} if $\exists A_1, A_2 \in \sub(A)$ such that $A_1 \neq A_2$, $\conc(A_1) = \conc(A_2)$, $A_1 \notin \sub(A_2)$ and $A_2 \notin \sub(A_1)$.
\end{definition}
\begin{example}\label{ex:redundancy}
Considering Example~\ref{ex:circularity-redundancy} in the light of
Definition~\ref{def:redundant:argument}, the two sub-arguments that define the redundancy of $E$ are 
$E_2 = (\{p\}, \{p\tTo r\}, r)$ and
$E_5 = (\{q\}, \{q\tTo r\}, r)$.
\end{example}
We say that arguments that are non-circular and non-redundant are \emph{regular} arguments since they are the kinds of argument that one encounters most often in the literature.
Clearly this is the same as saying:
\begin{definition}\label{def:regular:argument}
  Let $AT$ be an argumentation theory and $A \in \args{AT}$.
  We say that $A$ is \emph{regular} if $\nexists A_1, A_2 \in \sub(A)$ such that $A_1 \neq A_2$ and $\conc(A_1) = \conc(A_2)$.
\end{definition}
Now, to tie this back to the other notion of minimality, that of a minimal set of information from which a conclusion is derived, we need a notion of inference that works for \aspic.
 We start with a notion of closure.
Given an argumentation theory, we can define the closure of a set of propositions in the knowledge base under a set of rules of the theory.
\begin{definition}\label{def:closure}
Let $AT = \langle AS, \mc{K}\rangle$ be an argumentation theory, where $AS$ is the argumentation system $AS = \langle \lang, \overline{\phantom{\cdot}}, \Rules, n \rangle$.
We define the \emph{closure} of a set of propositions $P \subseteq \mc{K}$ under a set of rules $R \subseteq \Rules$ as $\closure(P)_{R}$, where:
\begin{enumerate}
	\item $P \subseteq \closure(P)_{R}$;
	\item if $p_1, \ldots, p_n \in \closure(P)_{R}$ and $p_1, \ldots, p_n \tTo p \in R$, then $p \in \closure(P)_{R}$; and
	\item $\nexists S \subset \closure(P)$ such that $S$ satisfies the previous conditions.
\end{enumerate} 
\end{definition}	
Based on the notion of closure, we can define a notion of inference from a set of propositions and rules of an argumentation theory.
\begin{definition}\label{def:inference}
Let $AT = \langle AS, \mc{K}\rangle$ be an argumentation theory, where $AS$ is the argumentation system $AS = \langle \lang, \overline{\phantom{\cdot}}, \Rules, n \rangle$.
	Given a set of propositions $P \subseteq \mc{K}$, a set of rules $R \subseteq \Rules$ and a proposition $p \in \mc{K}$, 
	we say that $p$ is inferred from $P$ and $R$, noted as $P \vdash_{R} p$, if $p \in \closure(P)_{R}$.
\end{definition}
Now, with this notion of inference, we can characterize \emph{minimal arguments}. These arguments are such that they have minimal (with respect to $\subseteq$) sets of grounds and rules that allow to infer their conclusion.
\begin{definition}\label{def:minimal:argument}
Let $AT = \langle AS, \mc{K}\rangle$ be an argumentation theory and $A \in \args{AT}$. We say that $A = (G, R, p)$ is a \emph{minimal argument} if $\nexists G'\subset G$ such that $G'\vdash_{R} p$ and $\nexists R'\subset R$  such that $G\vdash_{R'} p$.
\end{definition}
The following example illustrates the first condition in Definition~\ref{def:minimal:argument}.
\begin{example}\label{ex:minimality}
Let $AT = \langle AS, \mc{K}\rangle$ be an argumentation theory, where $AS = \langle \lang, \overline{\phantom{\cdot}}, \Rules, n \rangle$, $\Rules = \{d \tTo b; \mbox{$b \tTo c$}; \linebreak b, c \tTo a\}$ and $\mc{K} = \{b,d\}$.
From $AT$ we can construct the following arguments:
	\[
	\begin{array}{l}
	A_1 = [d]; A_2 = [A_1 \tTo b];  A_3 = [A_2 \tTo c];  A_4 = [b]; A = [A_4, A_3 \tTo a];\\
	B = [A_2, A_3 \tTo a];\\
	A_5 = [A_4 \tTo c]; C = [A_4, A_5 \tTo a]
	\end{array}
	\]
Here,  $A = (G, R, a)$, with $G = \{b, d\}$ and $R = \Rules$. In this case, $A$ is not minimal since $\exists G' \subset G$, with $G' = \{d\}$, such that $G' \vdash_R a$; moreover, $B = (G', R, a)$. 
On the other hand, argument $C$ is represented by the triple $(G'', R', a)$, with $G'' = \{b\}$ and $R' = \{b \tTo c; b, c \tTo a\}$.
In particular, argument $C$ is minimal. Furthermore, $B$ is also minimal since, even though $R' \subset R$, it is not the case that $G' \vdash_{R'} a$. 
\end{example}

It should be noted that, since the notion of minimality characterized in Definition~\ref{def:minimal:argument} explicitly accounts for the set of grounds of the arguments, this notion of minimality is different from those used in structured argumentation systems such as DeLP~\cite{garcia:simari:tplp}. Arguments in DeLP do not include the grounds: they are specified by a pair $(c, \Delta)$, where $\Delta$ is the set of rules used to derive the conclusion $c$. Then, the notion minimality in DeLP regards only the set of rules of the arguments. As a result, if we consider the arguments given in Example~\ref{ex:minimality}, argument $B$ would not be minimal in DeLP.

To illustrate the second condition of Definition~\ref{def:minimal:argument}, let us consider the situation depicted in Example~\ref{ex:circularity}. There we have arguments $A$ and $B$, which are both described by the triple $(G, R, a)$, with $G = \{a\}$ and $R = \{\mbox{$a \tTo c$}; c \tTo b; b \tTo a \}$. Also, there is argument $A_1 = (G', \emptyset, a)$, with $G' = \{a\}$. As a result, $\exists G' \subset G$ such that $G' \vdash_R a$ and therefore, arguments $A$ and $B$ are not minimal, in contrast with $A_1$.

Given the characterization of regular and minimal arguments, the following proposition shows that these notions are equivalent.
\begin{proposition}\label{prop:equiv:regular:minimal}
Let $AT = \langle AS, \mc{K}\rangle$ be an argumentation theory and $A = (G, R, p) \in \args{AT}$.
$A$ is a regular argument iff $A$ is a minimal argument.
\begin{proof}
The proof follows the same form as that of Proposition~\ref{prop:derivation}, being based around the three clauses of Definition~\ref{def:argument}.

Let us start with the if part. 
In the first clause of Definition~\ref{def:argument}, $p$ is a proposition in \mc{K}, $R$ is empty, and $G$ contains just $p$.
Clearly, in this case there is no $R'\subset R$, or $G' \subset G$ such that $G' \vdash_{R} p$ or $G \vdash_{R'} p$, so $A$ is minimal. It is also regular.
The second and third clauses in Definition~\ref{def:argument} define the recursive case.
Here, $A = (G, R, p)$ is an argument if $p$ is the conclusion of a rule, let us call it $r$ and there is an argument in \args{AT} for each of the premises of $r$.
$G$ is then the union of the grounds of all the arguments with conclusions that are premises of $r$, we will call this set of arguments $\textbf{Args}$, and $R$ is the union of all the rules for $\textbf{Args}$, call them \textbf{Rs}, plus $r$. 
If all the arguments in \textbf{Args} are minimal, then $A$ will be minimal, so long as $(i)$ adding $r$ does not introduce any non-minimality, and $(ii)$ the union of the grounds and the rules of the arguments in \textbf{Args} do not introduce any non-minimality. 
Let us consider case $(i)$.
For the addition of $r$ to introduce non-minimality, it must be the case that $(G, \textbf{Rs}, p)$ is an argument. In this case, $(G, \textbf{Rs}, p)$ will be a sub-argument of $A$ and thus, by Definition~\ref{def:circular:argument} $A$ is circular, contradicting the hypothesis that it is a regular argument.
Let us now consider case $(ii)$. 
Here, in order for $A$ not to be minimal, there have to be  minimal arguments $(G_1, R_1, p_1), \ldots, (G_n, R_n, p_n)$ in $\textbf{Args}$ such that $p_1, \ldots, p_n$ are the premises in $r$ and $A = (\bigcup_{i=1}^{n}G_i, \bigcup_{i=1}^{n}R_i\cup\{r\}, p)$ is not minimal.
Because we are taking the unions, no duplication can be introduced.
Since $G_1, \ldots, G_n$ are just sets of propositions, their union cannot be the cause of any non-minimality, and we know from case $(i)$ that any non-minimality is not due to $r$. 
So if any non-minimality is introduced, it is in $\bigcup_{i=1}^{n}R_i$.
Since Proposition~\ref{prop:derivation} tells us that every rule in $R_i$ must be used in deriving $p_i$, the only way that $\bigcup_{i=1}^{n}R_i$ can make $ A$ non-minimal is if there is some rule in $R_j$ which allows the derivation of the same conclusion as a rule in $R_k$ $($with $1 \leq j,k \leq n$, and $j\neq k$). 
If this is the case, $A = (G, R, p)$ would have two distinct sub-arguments with the same conclusion, where one is not a sub-argument of the other, and hence, by Definition~\ref{def:redundant:argument} be redundant, contradicting the hypothesis that $A$ is regular.

Let us now address the only if part.
In the first clause of Definition~\ref{def:argument}, $p$ is a proposition in \mc{K}, $R$ is empty, and $G$ contains just $p$. Clearly, in this case $A$ is regular since $A$ is the only sub-argument of $A$; thus, there exist no distinct sub-arguments of $A$ with the same conclusion.
$A$ is also minimal.
The second and third clauses in Definition~\ref{def:argument} define the recursive case. 
Here, $A = (G, R, p)$ is an argument if $p$ is the conclusion of a rule, let us call it $r$ and there is an argument in \args{AT} for each of the premises of $r$.
$G$ is then the union of the grounds of all the arguments with conclusions that are premises of $r$, and $R$ is the union of all the rules for those arguments, plus $r$. 
Since by hypothesis $A = (G, R, p)$ is minimal, it must be the case that $\nexists G' \subset G$ such that $G' \vdash_R p$ and $\nexists R' \subset R$ such that $G \vdash_{R'} p$.
Suppose by contradiction that $A$ is not regular. Hence, there should exist two distinct sub-arguments $A_1 = (G_1, R_1, p')$ and $A_2 = (G_2, R_2, p')$ of $A$ such that $G_1 \neq G_2$, $R_1 \neq R_2$ or both. 
However, this would imply that $\exists G' \subset G$ $($with $G' = (G \backslash G_1) \cup G_2$ or $G' = (G \backslash G_2) \cup G_1)$ or $\exists R' \subset R$ $($with $R' = (R \backslash R_1) \cup R_2$ or $R' = (R \backslash R_2) \cup R_1)$ such that $G' \vdash_R p$ and $\nexists R' \subset R$ such that $G \vdash_{R'} p$, contradicting the hypothesis that $A$ is minimal.
\end{proof}
\end{proposition}
The following example illustrates the relationship between regular and minimal arguments.
\begin{example}
Let us consider the arguments from Example~\ref{ex:minimality}, where it was shown that $B$ and $C$ are minimal arguments, whereas $A$ is not.
Here, we have that $B$ is also regular, since it has no pair of sub-arguments with the same conclusion. Specifically, $\sub(B) = \{B, A_2, A_3, A_1\}$, and $\conc(B) = a$, $\conc(A_2) = b$, $\conc(A_3) = c$, $\conc(A_1) = d$.
Similarly, $C$ is also regular since $\sub(C) = \{C, A_4, A_5\}$, where $\conc(C) = a$, $\conc(A_4) = b$ and $\conc(A_5) = c$.
In contrast, if we consider argument $A$, which was shown to be non-minimal in Example~\ref{ex:minimality}, we have  $\sub(A) = \{A, A_4, A_3, A_2, A_1\}$ where, in particular, $\conc(A_4) = b$ and $\conc(A_2) = b$; therefore, $A$ is not a regular argument.	
	
On the other hand, if we consider the arguments from Example~\ref{ex:circularity-redundancy}, it was shown in Examples~\ref{ex:circularity-twoforms} and~\ref{ex:redundancy} that $A$, $C$ and $E$ are not regular arguments (the first two by being circular and the last one by being redundant).
Then, if we look at the minimality of these arguments, we have that $A = (G_a, R_a, t)$, with $G_a = \{p, q, r\}$, $R_a = \{p, q \tTo s; s \tTo q; q, r \tTo t\}$, and $\exists G'_a = \{r\}$, $\exists R'_a = \{r \tTo t\}$ such that $G'_a \vdash_{R_a} t$ and $G_a \vdash_{R'_a} t$; hence, $A$ is not a minimal argument.
In the case of $C = (G_c, R_c, r)$, with  $G_c = \{p, q\}$ and $R_c = \linebreak \{p, q \tTo r; r \tTo s; s \tTo t; t \tTo r\}$, we have that $\exists R'_c = \{p, q \tTo r\}$ such that $G_c \vdash_{R'_c} r$ and therefore, $C$ is not minimal.
Finally, given $E = (G_e, R_e, t)$, with $G_e = \{p, q\}$ and $R_e = \{p \tTo r; r \tTo s; q \tTo r; r, s \tTo t\}$, it is the case that $\exists G'_e = \{p\}$, $\exists G''_e = \{q\}$, $\exists R'_e = \{p \tTo r; r \tTo s; r, s \tTo t\}$, $\exists R''_e = \{r \tTo s; q \tTo r; r, s \tTo t\}$ such that $G'_e \vdash_{R_e} t$, $G''_e \vdash_{R_e} t$, $G_e \vdash_{R'_e} t$ and $G_e \vdash_{R''_e} t$; thus, $E$ is not a minimal argument.
\end{example}
Let us consider another example regarding minimal and non-minimal arguments.
\begin{example}\label{ex:same:conclusion}
Consider that we have an argumentation system $AS_{\ref{ex:same:conclusion}} = \langle \lang_{\ref{ex:same:conclusion}}, \overline{\phantom{\cdot}}, \Rules_{\ref{ex:same:conclusion}}, n \rangle$, where
$\lang_{\ref{ex:same:conclusion}} = \{p, q, r, \neg p, \linebreak \neg q, \neg r\}$ 
and $\Rules_{\ref{ex:same:conclusion}} = \{p \tTo q;  q \tTo r\}$. 
By adding the knowledge base $\mc{K}_{\ref{ex:same:conclusion}} = \{p, q\}$ we obtain the argumentation theory $AT_{\ref{ex:same:conclusion}} = \langle AS_{\ref{ex:same:conclusion}}, \mc{K}_{\ref{ex:same:conclusion}}\rangle$, from which we can construct the following arguments:
\[
\begin{array}{lll}
H_1 = [p]; & H_2 = [H_1 \tTo q]; & H = [H_2 \tTo r];\\
I_1 = [q]; & I = [I_1 \tTo r]
\end{array}
\]
such that $H = (\{p\}, \{p \tTo q; q \tTo r\}, r)$ and 
$I = (\{q\}, \{q \tTo r\}, r)$.	
\end{example}
Even though arguments $H$ and $I$ in Example~\ref{ex:same:conclusion} have the same conclusion and use the same rule to draw that conclusion, they are both minimal. This is because an argument is minimal according to Definition~\ref{def:minimal:argument} if there is no argument for the same conclusion with a smaller set of both premises (grounds) and rules.
  Thus it is possible to have two minimal arguments for the same conclusion, one of which uses a subset of the rules that the other does, so long as the premises are disjoint.
  Similarly, we could have two minimal arguments for the same conclusion, one of which uses a subset of the premises that the other does, so long as the rules are disjoint.

On the other hand, the situation depicted in Example~\ref{ex:same:conclusion} relates to the one involving arguments $C_3 = (\{p, q\}, \{p, q \tTo r\}, r)$ and $C = (\{p, q\}, \{p, q \tTo r; r \tTo s; s \tTo t; t \tTo r\},r)$ in Example~\ref{ex:circularity-redundancy}. However, even though $C_3$ and $C$ have the same conclusion, differently from $H$ and $I$, they are such that one is a sub-argument of the other (specifically, $C_3$ is a sub-argument of $C$) and, as a result, $C$ is not regular nor minimal.

Finally, it should be noted that, since \aspic\ arguments are not required to be minimal in the sense of Definition~\ref{def:minimal:argument}, it can be the case that two different arguments $A$ and $B$ have the same description as a triple $(G,R,p)$, as occurred in Example~\ref{ex:circularity}. 
However, as shown by the following proposition, that cannot be the case when considering minimal arguments.

\begin{proposition}\label{prop:minimal:description}
Let $AT = \langle AS, \mc{K}\rangle$ be an argumentation theory and $A = (G, R, p) \in \args{AT}$.
If $A$ is a minimal argument, then $\nexists B \in \args{T}$ such that $B \neq A$ and $B = (G, R, p)$.
\begin{proof}
  Suppose that $A = (G, R, p)$ is a minimal argument and $\exists B \in \args{T}$ such that $B \neq A$ and $B = (G, R, p)$.
  By Proposition~\ref{prop:relevance:grounds-rules}, every element in the grounds $G$ and every rule in $R$ is used in the derivation of $A$'s and $B$'s conclusion $p$.
  Furthermore, since by hypothesis $A$ is minimal, by Definition~\ref{def:minimal:argument} it is the case that $\nexists G' \subset G$, $\nexists R' \subset R$ such that $G' \vdash_R p$ or $G \vdash_{R'} p$.
  If $B \neq A$, then it must be the case that the difference between them is on the number of times they use the rules in $R$.
  Since by hypothesis $A$ is minimal, there must be a rule $r = c_1, \ldots, c_n \tTo c \in R$ that is used more times in $B$ than in $A$.
  Now consider the derivation of $A$ and $B$.
    From what we have said so far, these must be largely the same, so we can think of them starting from the same set of premises and applying rules, one by one.
  Thinking of the two arguments like this, side by side, so to speak, since $B$ uses some $r$ more than $A$, then
  at some stage $B$ uses the rule $r$ to derive $c$, whereas $r$ is not used in $A$ at that point.
  Hence, since $c$ is needed at that point to derive $A$'s conclusion $p$, there must be an alternative derivation for $c$ in $A$, which does not require the use of $r$.
  However, this would imply that there exists a rule $r' \in R$ such that $r' = c'_1, \ldots, c'_m \tTo c$ or $c \in G$, contradicting the hypothesis that $A$ is minimal.
  As a result, if $A$ is a minimal argument, then $\nexists B \in \args{T}$ such that $B \neq A$ and $B = (G, R, p)$. 
\end{proof}
\end{proposition}

We end by noting that even though $H$ and $I$, above, are both minimal in the sense of Definition~\ref{def:minimal:argument}, $I$ is in some sense ``more minimal'' than $G$ since the sets of premises of both arguments are the same size, while $I$ has a smaller set of rules.
  This suggests that further forms of minimality may be worth investigating.

\section{Conclusion}
\label{sec:discuss}

In this report we have studied the notion of minimality of arguments in the context of \aspic.
We have considered two forms of minimality.
The former corresponds to the native minimality of \aspic, which implies that arguments do not include irrelevant grounds or rules.
We have noted that, under the native form of minimality, redundant and circular arguments may be obtained. 
Although there is nothing inherently wrong with circular and redundant arguments, in some cases it may be helpful to work with arguments that satisfy a stronger form of minimality. 
A stronger form of minimality is satisfied by what we have identified as \emph{regular} arguments, since these are the arguments that one encounters most often in the literature of argumentation. 
Specifically, regular arguments make use of a minimal set of grounds and rules when deriving their conclusion.
It should be noted that an argument $A$ satisfying the stronger form of minimality uses the same grounds and rules for deriving a proposition $c$ at every step in which $c$ is required in the derivation of $A$'s conclusion. 
Furthermore, we have shown that regular arguments, satisfying the stronger form of minimality, can be unequivocally described by a triple $(G, R, p)$, distinguishing their grounds, rules and conclusion.
In contrast, that is not the case for arguments complying only with \aspic's native form of minimality. 
Finally, in the future we are interested in studying the relationship between the strong form of minimality characterized here and that considered in other structured argumentation systems.


\bibliographystyle{plain}

\end{document}